\documentclass[10pt, twocolumn, letterpaper]{article}

\usepackage{cvpr}
\usepackage{times}
\usepackage{epsfig}
\usepackage{graphicx}
\usepackage{amsmath}
\usepackage{amssymb}
\usepackage{cuted}

\usepackage[breaklinks=true,bookmarks=false]{hyperref}

\cvprfinalcopy % *** Uncomment this line for the final submission

\def\cvprPaperID{2719} % *** Enter the CVPR Paper ID here

\ifcvprfinal\fi
\begin{document}

%%%%%%%%% TITLE
%\title{\LaTeX\ Tinkering Under The Hood: Pictorial Languages with Applications to Interactive-Zero Shot Learnings}
%\title{Tinkering Under The Hood: Building Explainable Models from Deep CNNs with Interactive Methods}
%\title{Tinkering Under the Hood: Using CNN Pictorial Language Representations to Build Explainable Models for Novel Visual Concepts}
\title{Tinkering Under the Hood: Interactive Zero-Shot Learning with Net Surgery}

\author{Vivek Krishnan \qquad Deva Ramanan\\
Carnegie Mellon University\\
Pittsburgh, PA 15213\\
{\tt\small \{vrkrishn,deva\}@cs.cmu.edu}
}

\maketitle

\begin{strip}
     \centering\noindent
     \includegraphics[width=.9\linewidth]{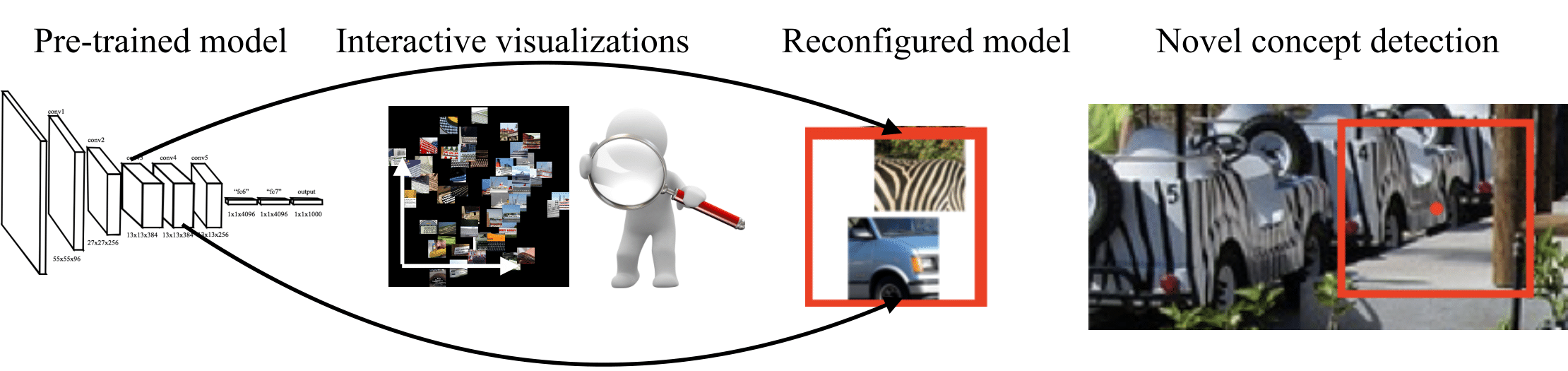}
    \begin{quote}
		We introduce a pictorial language for visualizing, understanding, and {\em re-wiring} internal representations learned by CNNs to detect never-before-seen concepts, such as a ``striped car".
	\end{quote}
\end{strip}

%%%%%%%%% ABSTRACT
\begin{abstract}
%    \vspace{-0.3cm}
    We consider the task of visual net surgery, in which a CNN can be reconfigured without extra data to recognize novel concepts that may be omitted from the training set. While most prior work make use of linguistic cues for such "zero-shot" learning, we do so by using a \textit{pictorial language} representation of the training set, implicitly learned by a CNN, to generalize to new classes. To this end, we introduce a set of visualization techniques that better reveal the activation patterns and relations between groups of CNN filters. We next demonstrate that knowledge of pictorial languages can be used to rewire certain CNN neurons into a part model, which we call a pictorial language classifier. We demonstrate the robustness of simple PLCs by applying them in a weakly supervised manner: labeling unlabeled concepts for visual classes present in the training data. Specifically we show that a PLC built on top of a CNN trained for ImageNet classification can localize humans in Graz-02 and determine the pose of birds in PASCAL-VOC without extra labeled data or additional training. We then apply PLCs in an interactive zero-shot manner, demonstrating that pictorial languages are expressive enough to detect a set of novel visual classes in MS-COCO.
    
\end{abstract}

%\begin{abstract}
%   We consider the task of visual zero-shot learning, in which a system must learn to recognize concepts omitted from the training set. While most prior work make use of linguistic cues to do this, we do so by using a \textit{pictorial language} representation of the training set, implicitly learned by a CNN, to generalize to new classes. We first demonstrate the robustness of pictorial language classifiers (PLCs) by applying them in a weakly supervised manner: labeling unlabeled concepts for visual classes present in the training data. Specifically we show that a PLC built on top of a CNN trained for ImageNet classification can localize humans in Graz-02 and determine the pose of birds in PASCAL-VOC without extra labeled data or additional training. We then apply PLCs in an interactive zero-shot manner, demonstrating that pictorial languages are expressive enough to detect a set of visual classes in MSCOCO that never appear in the ImageNet training set.
%\end{abstract}

%%%%%%%%% BODY TEXT
\section{Introduction}
Convolutional Neural Networks (CNNs) have revolutionized modern vision pipelines by replacing hand-crafted features with large set of trainable parameters that learn complex image representations from data. However, these models notoriously require a substantial amount of annotated training data to learn a feature hierarchy that generalizes well. As a result, subsequent performance gains in various visual tasks have been limited by the availability of large datasets such as ImageNet~\cite{deng2009imagenet}%and Pascal VOC~\cite{everingham2010pascal}

\textbf{Explainable AI:}
 A growing reactionairy response to deep networks is that their opaqueness makes them difficult to use when more than a simple prediction is required. If a network fails, why? How will a network behave on never-before-seen data? To answer such questions, there is a renewed interest in so-called {\em explainable AI}\footnote{\url{http://www.darpa.mil/program/explainable-artificial-intelligence}}. One goal in this reinvigorated field is the development of machine learning systems that are designed to be more interpretable and explanatory.
 %'DARPA has issued a challenge to create "new machine learning methods to produce more explainable models and combine them with explanation techniques." In the presentation, DARPA specifically calls for new methods of training that provide an explanation interface for users to justify model outputs.
 
{\bf Interpreting Networks:} Rather than redesigning a deep network architecture to become more interpretable, a complementary approach is to {\em post-hoc} add interpretable semantics to an existing model. Deep hierarchical models often dramatically outperform manually-designed hierarchies (classically based upon edges, parts, objects~\cite{marr1982vision}), so it would be valuable to understand what non-obvious concepts are being learned by such high-performing deep networks. An illustrative example might be post-hoc interpretation of the AlphaGo network~\cite{silver2016mastering}; while an automated Go player itself is certainly interesting, what maybe more so is understanding the non-obvious heuristics it learns to employ.

{\bf Model reconfigurability:}
Evaluating post-hoc interpretability is challenging. In this work, we propose an approach based on reconfigurability; that is,  we quantitatively evaluate the interpretability of a model by the {\em ability to interactively re-purpose it for never-before-seen tasks}. A practical issue with virtually any real-world deployment of machine-learned models is the ability to adapt to novel conditions. Such a problem formulation is often cast as domain adaption or transfer learning. Indeed, virtually all state-of-the-art vision systems make use of transfer learning via models pre-trained on ImageNet~\cite{girshick2014rich,%razavian2014cnn,
yosinski2014transferable}. A related but distinct formalism is that of domain adaptation, where a pre-trained model is adapted to a novel domain with different statistics from test data~\cite{ben2007analysis,saenko2010adapting}  or entirely different modalities~\cite{glorot2011domain,tzeng2015simultaneous}. However, in most systems, some amount of target data for the task of interest is needed to provide a gradient-based signal for learning. %(even when fine-tuning).\\

\begin{figure*}[t!]
    \centering
    \includegraphics[width=\textwidth]{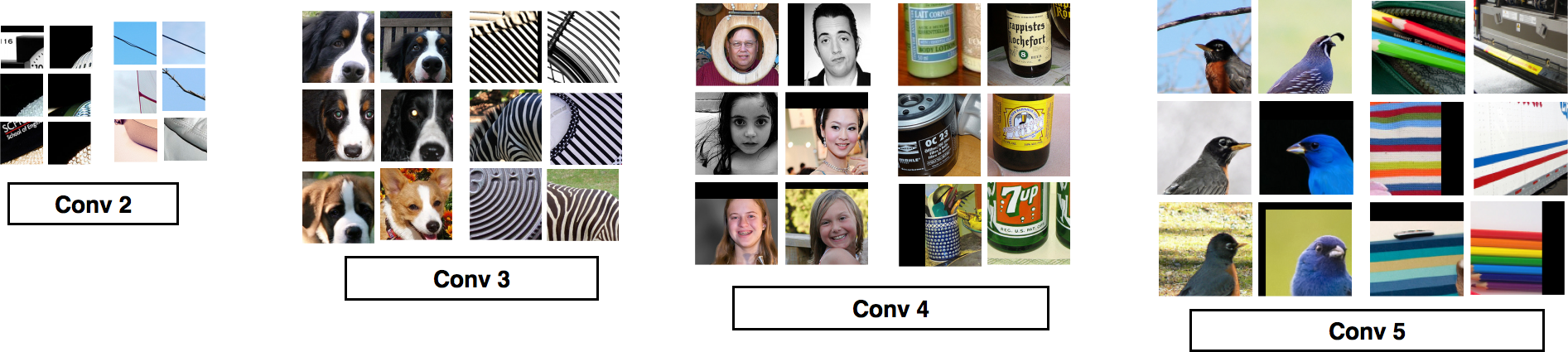}
    % \vspace{-10pt}
    \caption{Visualization of image features % in the VGG\_CNN\_S model.
    from AlexNet. For layers 2-6, we show the top 6 activations taken from the ILSVRC 2012 \cite{russakovsky2015imagenet} validation set. The size of the image patch corresponds to the theoretical RF of the neuron as computed in \cite{long2014convnets,girshick2015fast}. We see that (1) a CNN trained for object classification learns a rich, hierarchical set of features, ranging from textures to complex invariances such as pose. (2) Neurons can learn to identify and localize unlabeled concepts (such as face). (3) These images of faces belong to a wide range of classes including Helmet, French Horn, Suit, Bathtub, and Park Bench. Even though there is no category on ILSVRC 2012 for generic human detection, the CNN has automatically determined that it is useful to reason about human faces as intermediate concepts. }
    % \vspace{-10pt}
    \label{fig:viz}
\end{figure*}

 {\bf Interactive zero-shot learning:}
The most difficult instantiation may arguably be the zero-shot setting, where no data for the novel task is provided. In this setting, even specifying the new task is difficult. One traditionally makes use of an external semantic knowledge base to do so~\cite{NIPS2009_3650}. For example, an existing lexical database may define an ``elund" as a ``striped animal with horns". This knowledge can be combined with existing striped and horn models to yield a visual model of a never-before-seen object. The vast majority of past work makes use of linguistic semantics, mined from text corpora~\cite{%frome2013devise,
fu2015zero} or lexical knowledge bases such as Wordnet~\cite{miller1995wordnet}. Instead, we propose to use a human as the external knowledge base, yielding a novel form of {\em interactive zero-shot learning}. The output of such a process is a model where spatial logic and combination of parts follows directly from human logic.

{\bf Pictography:} Rather than relying on linguistics, our approach encodes semantic concepts pictorially in a manner that is intuitive for a human user. Cognitive psychology suggests that many concepts in one's associative memory may be better represented through visual imagery rather than linguistic tokens~\cite{paivio1969mental}. Similarly, it is well-known in developmental psychology that children learn spatial concepts visually rather than linguistically~\cite{%marzano2010representing,
bowerman1996origins}. %Building on such obserations, prior work on "zero-shot" image retrieval has shown that sketch-based interfaces can be more effective than textual input for certain queries~\cite{antol2014zero}. 
Inspired by these observations, we demonstrate that CNN visualizations can be used to define a {\em pictorial language} that can be used to interactively reconfigure networks for novel tasks.

{\bf Visualizations:} Our pictorial approach builds on a body of work that aims to visualize and understand internal representations in CNNs. Common visualization techniques include reconstructing image filters (by deconvolution~\cite{zeiler2014visualizing}), inverting feature transformations~\cite{mahendran2015understanding}, or computing the set of image patches that maximally activate particular neurons~\cite{girshick2014rich,zhou2014object}. Instead, we interpret local neighborhoods of neural activations as {\em embeddings}, and use classic techniques for visualizing low-dimensional projections such as tSNE ~\cite{van2008visualizing,maaten2009learning}. %This allows users to extract subtle visual cues from individual neurons by "drawing" linear classifiers in the low-dimensional space. 
%Such subtle concepts can be composed together in a 
Importantly, our visualizations allow users to discover new pictorial concepts that cannot be represented by any single neuron. We then introduce a simple visual grammar for composing these pictorial concepts together to describe never-before-seen objects (such as a ``striped car"). Our grammars allow a user to build zero-shot classifiers composed out of visual tokens that may not have a natural linguistic ``name". Our grammars can be seen as spatial part models~\cite{girshick2015deformable,zhu2007stochastic} where part responses are equivalent to interactively-defined functions of activations (computed through our embedding visualizations).

{\bf Overview:} 
We begin with a novel techniques for visualizing neurons in deep networks based on embeddings (Sec.~\ref{sec:viz}). We describe a pictorial grammar that uses such visualizations to repurpose existing deep networks for novel tasks (Sec.~\ref{sec:model}). We use these techniques in Sec.~\ref{sec:exp} to repurpose a network trained for image {\em classification} (~\cite{krizhevsky2012imagenet}) for object {\em detection}, including never-before-seen objects.

\section{Visualizing Activations}
\label{sec:viz}
 
In this section, we describe our approach for visualizing and understanding CNNs. Though our methods directly apply to any existing network, we illustrate them on our running example of AlexNet. Our methods fundamentally require both a pre-trained CNN and a collection of validation images with which to process the CNN and diagnose its behaviour. We use the ILSVRC 2012 validation set of 50,000 images~\cite{russakovsky2015imagenet}. Importantly, this set is distinct from the test images we use for evaluation.
%explore the image representations expressed throughout a CNN trained for object classification on ImageNet. We use visualization and analytical techniques to find specific neurons that correspond to interesting parts or features. Finally, we discuss methods to convert these neurons into part models for our template-based object detector described in Sec.~\ref{sec:model}.

%\subsection{Visualizing neurons}

\begin{figure}[!ht]
\centering
    \includegraphics[width=\linewidth]{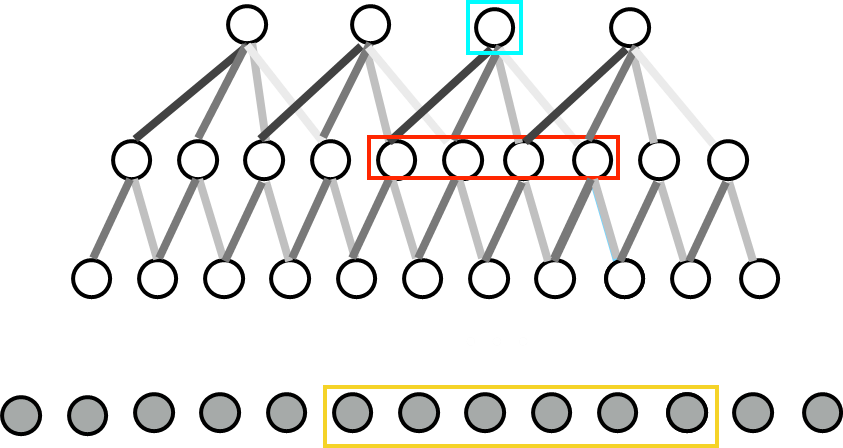}
   \caption{Our retinotopic view of CNNs. Here \textit{a} is the activation of particular CNN neuron (cyan). We embed this neuron as a feature vector $w \cdot x + b$, where $x$ is a set of input activations (red). The receptive field in the original image corresponding to neuron $a$ is drawn in yellow.}
\label{fig:net}
\end{figure}

\subsection{Retinotopic embeddings}

\label{pca}
%Each neuron in a CNN is optimized to look for a specific image feature, with higher neuron activations corresponding to greater confidence of positive feature detection. Therefore, visualization of the image features learned by a neuron reduces to retrieving the image patches that maximally activate it. In Fig \ref{fig:viz_Fig} we apply this technique to a subset of ImageNet CNN neurons taken from different layers. From these visualizations, we see the CNNs learn a rich, hierarchical feature representation that can identify textures and shapes, localize object parts, and can be invariant to different object poses.

The heart of our approach requires the ability to interpret the semantics of particular neurons, where neurons correspond to individual activations computed throughout the layers of a CNN. Note that it is not clear that this is even possible; indeed, the well-known ``grandmother neuron" hypothesis that individual neurons represent semantic concepts (such as one's grandmother) seems at odds with notion that deep networks are distributed (rather than localized) representations~\cite{bengio2009learning}. We will shortly offer a solution to this apparent contradiction.

Following established convention, we will denote activations by their layer, such as {\tt conv3}, {\tt conv4}, etc. While numerous techniques have been proposed for understanding the semantics of such neurons, our starting point is a surprisingly simple strategy: characterize a neuron by the {\em set} of $N$ image patches that maximally activate that particular neuron. Such an approach was popularized by~\cite{girshick2014rich,zhou2014object}. We show examples for $N=6$ for a variety of neurons across different layers in Fig.~\ref{fig:viz}; one can readily identify neurons corresponding to different textures, parts, and objects. Neurons in lower layers are characterized by smaller \textit{receptive fields} (defined as the set of pixel values that will affect a given neural activation). Neurons in deeper layers tend to be invariant to various factors such as pose, but perhaps surprisingly, strong semantics appear even at shallower layers (e.g., the dog-face neuron in {\tt conv3}).

\begin{figure}[t]
\centering
    \includegraphics[width=\linewidth]{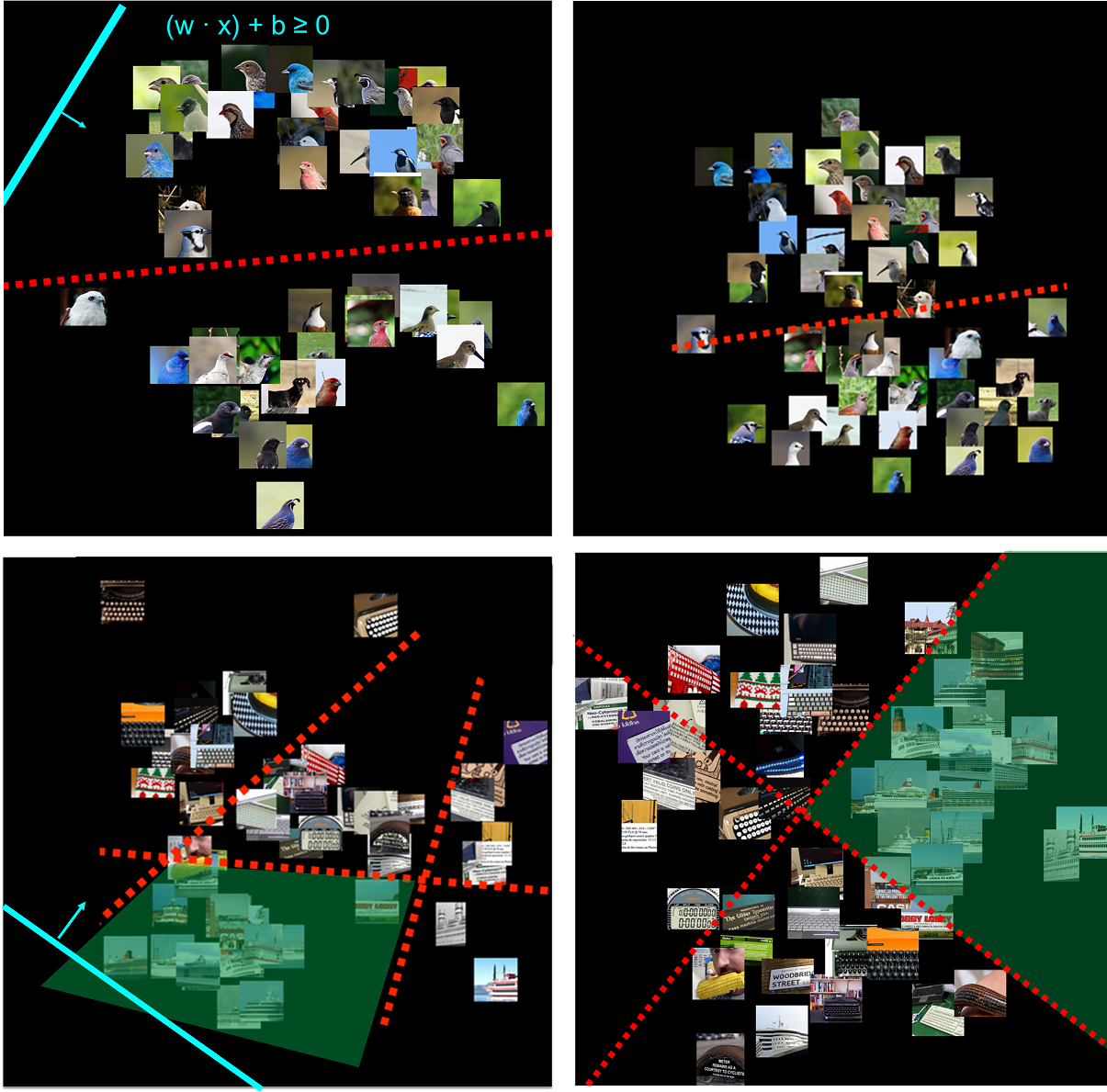}
%     \vspace{-10pt}
   \caption{We visualize pairs of (PCA,TSNE) embedding of two \texttt{conv5} neurons, along with user-drawn linear boundaries (dotted red lines. For the PCA visualizations (left), we show a projection of the original filter weight into 2D space using the PCA basis (blue line). Because the t-SNE projections (right) are non-linear, we cannot simply project the filter weight in such a fashion. %For PCA, the x-axis is defined as the maximum-variance direction. 
   %For PCA, we also visualize the direction $w$ (the yellow arrow) that would maximally excite the given neuron, %$a$,  %= \max(0,w\cdot x+b)$ 
   %rather than the linear threshold boundary which lies outside the image.
   %Given the input activations corresponding to $N=25$ patches, we project them into 2 dimensions using PCA (such that x-axis is maximum variance). Because all these patches produce positive activations, we do not visualize the linear threshold corresponding to $a$ (since it would lie outside the image) but visualize the direction $w$ ({\bf the yellow arrow}) that would maximally activate this neuron.
   %The top 25 images are embedded in 2d space using this projection. The decision boundary can be used to extract pose information from neurons that learn to differentiate between different object poses
   For the \textbf{top} pair, we clearly observe a linear separator in the embeddings between bird heads with different orientations. The \textbf{bottom} illustrates a grouping of texture patterns that may not be easily described linguistically, but appear to loosely correspond to keyboard, cruise ship, and text. We have highlighted the region corresponding to cruise ship as an example.
   }
%    \vspace{-10pt}
\label{fig:pca}
\end{figure}

{\bf Embeddings:} Because neurons at higher layers tend to be more invariant, we would like to understand the {\em variation} in the appearance of image patches that activate a particular neuron. Recall that a neuron activation $a$ is computed from previous layer responses with a (convolutional) linear dot product with weights $w$ and bias $b$, followed by a (ReLU) nonlinearity:
\begin{align}
    a = \max(0,w \cdot x + b), \label{eq:activate}
\end{align}
\noindent where $x$ corresponds to a local neighborhood of activations from the previous layer. For concreteness, if $a$ was from {\tt conv3}, $x$ is of dimensionality $3 \times 3 \times 256$. We propose to view $x \in \mathcal{X}$ as a point in a $\mathcal{X}=R^{2304}$ dimensional embedding space, where the activation $a$ is given by a linear threshold function in $\mathcal{X}$. We posit that this embedding space is a rich characterization of the invariances learned by this particular neuron - e.g., if neuron $a$ corresponds to a view-invariant object, then different points in $\mathcal{X}$ may correspond to different viewpoints of that object.  We use the embedding space  $\mathcal{X}$ to visualize image patches of size given by the receptive feild of neuron $a$ (Fig.~\ref{fig:net} and Fig.~\ref{fig:pca}).

{\bf Past work:} The idea of viewing neural activations as embeddings is certainly not new. Training deep nets to predict embedding coordinates is a well-studied task~\cite{hinton2006reducing}. Moreoever, it is widely known that one can interpret the final (FC7) activation layer as an embedding, even for nets originally trained for classification~\cite{devlin2015exploring}. Our key insight simply extends this observation to convolutional layers, which we demonstrate requires examining {\em local neighborhoods} of activations~\eqref{eq:activate}. This observation might offer a solution to the localized (grandmother) versus distributed debate: interpretable semantics might reside in localized neighborhoods of activations rather than any individual one. Or in other terms, we should {\bf not} conclude that a network cannot understand ``wheels" simply because we cannot find any single neuron that responds to ``wheels": rather, this knowledge may be encoded in a distributed fashion across a local neighborhood of neurons. Because such neighborhoods share similar receptive fields (due to the convolutional structure of weights), we refer to such embeddings as {\em retinotopic embeddings}.
%Hence we offer an intermediate option regarding the grandmother vs distributed hypothesis; semantic concepts are represented by local retinotopic neighborhoods of neural activations. 
This characterization of deep nets as learning progressively invariant embeddings may agree with perspectives of such models as ``disentangling" factors of variation from one layer to the next~\cite{bengio2009learning}.

{\bf  Dimensionality reduction:} We use the above observation to visualize $N$ high-scoring patches for a particular neuron. By viewing each patch as a point in $\mathcal{X}$, we can visualize a collection using low-dimensional projections (Fig.~\ref{fig:pca}). T-SNE~\cite{van2008visualizing} is a popular technique for visualizing {\em nonlinear} projections that preserves neighborhood structure in the high-dimensional space. We also experimented with simpler {\em linear} projections obtained through PCA, which also allows us to visualize neuron $a$ as a 2D line given by the linear threshold function $(w,b)$. Formally speaking, let us write the 2-dimensional projected coordinates of a point $x \in \mathcal{X}$ as $$(c_1,c_2) = (v_1 \cdot x, v_2 \cdot x) \quad \text{[PCA]},$$ where $(v_1,v_2)$ are projection vectors given by PCA. Given that we can approximately reconstruct $x$ from its projection with $x \approx c_1 v_1 + c_2v_2$, the linear threshold function from \eqref{eq:activate} can be written as $$wx + b \approx c_1 v_1 \cdot w + c_2 v_2 \cdot w + b,$$ or a 2D line with a normal vector $(v_1 \cdot w, v_2 \cdot w)$ and offset $b$. This line is visualized in blue in Fig.~\ref{fig:pca}.

{\bf Activations as 1-D embeddings:} It is worth noting that the {\em un}thresholded activation value $a' = w \cdot x + b$ is itself a one-dimensional linear projection of the embedding space $\mathcal{X}$. Hence prior work that analyzes networks by plotting the top $N$ scoring patches that activate any neuron~\cite{girshick2014rich,zhou2014object} can also be thought to employ low-dimensional (or rather, one-dimensional) embeddings. By projecting to more than a single dimension, our approach captures much richer semantics such as pose and categorical variation (Fig.~\ref{fig:pca}).

\textbf{User-drawn concepts (PCA):} Our embedding visualizations can be used to define {\em new} linear threshold boundaries, corresponding to user-defined visual concepts. In the case of PCA-embeddings, these user-drawn linear thresholds can be back projected to define new filters and biases. Formally, given a user-drawn line with normal vector $(\alpha_1,\alpha_2)$ and offset $\beta$, the corresponding high-dimensional filter producing the same response is given by 
\begin{align}
    \alpha_1 c_1 + \alpha_2 c_2 + \beta &= \alpha_1 v_1 \cdot x + \alpha_2 v_2 \cdot x + \beta \nonumber\\
&= w' \cdot x + b', \nonumber
\end{align}
where $w' = \alpha_1 v_1 + \alpha_2 v_2$ and $b' = \beta$. We will show examples of zero-shot models built with user-defined filters in our experimental results.

\begin{figure*}[t!]
\centering
   \includegraphics[width=\textwidth]{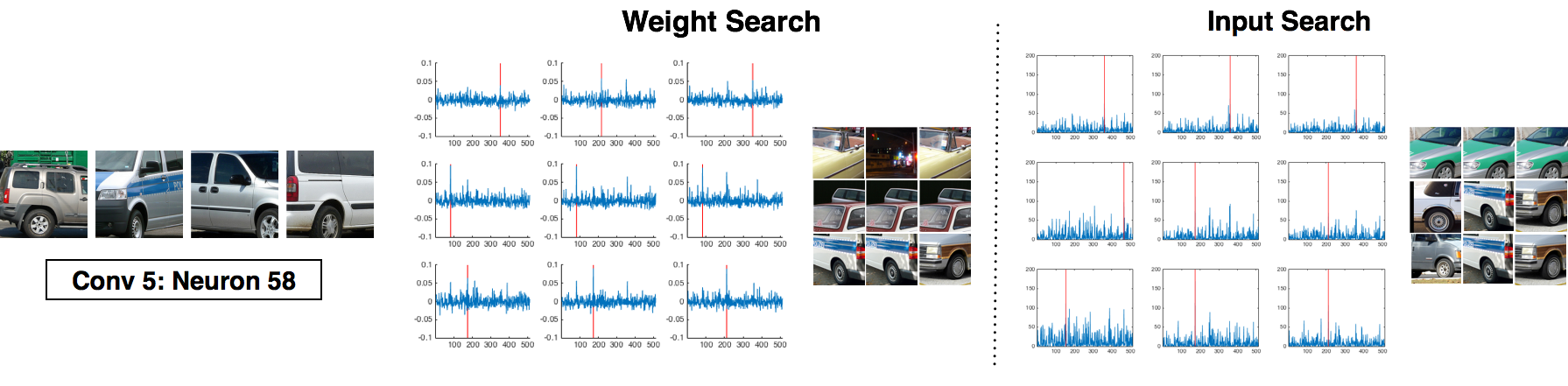}
   \caption{ We illustrate two search methods based on neuron ``58" from {\tt conv5}, which fires on images of vehicles. We provide the 4 image patches in the ILSVRC validation set that maximally activate the neuron (left). Weight Search ({\bf left}) consists of finding the maximum weight for each spatial location in the filter. We plot the learned weight values for each filter and indicate the maximum filter weight with a red line. Input Co-occurence Search ({\bf right}) consists of averaging the input features over the top-100 activations of the neuron. Here we plot the averaged input for each filter and highlight the maximum with a red line. For each search method, we also show the top image at each spatial location corresponding to the retrieved {\tt conv4} filters.  } 
\label{fig:search}
\end{figure*}

{\bf User-drawn concepts (TSNE):} Reconstructing the filter from a TSNE embedding is much more difficult. TSNE does not explicitly compute an embedding function, but rather directly outputs an embedding of a fixed set of input points (implying that the embedding cannot be applied to ``out-of-sample" points). P-TSNE~\cite{maaten2009learning} is a parametric extension that essentially trains a feedforward neural net to predict the TSNE embedding obtained for a fixed set of inputs - e.g., $$(c_1,c_2) = (f_1(x),f_2(x)) \quad \text{[TNSE]}.$$ Once this function is learned, it can be applied to new ``out-of-sample" patches. Because the function is nonlinear, halfspaces in the embedded space do not correspond to simple linear filters in $\mathcal{X}$. However, one can still ``implement" a user-drawn halfspace by explicitly computing the feedforward embedding on any query patch, and linearly scoring the resulting 2D projection: $$a_{TSNE} = \max(\alpha_1 f_1(x) + \alpha_2 f_2(x) + \beta,0).$$ Note that this same approach could also apply for PCA; instead of reconstructing the original filter, one could explicitly computes linear thresholds in the low-dimensional projected space $$a_{PCA} = \max(\alpha_1 v_1 \cdot x + \alpha_2 v_2 \cdot x + \beta,0),$$ which might be faster if processing multiple user-defined concepts obtained from the same embedding.

%When we back-project a decision boundary into the input feature space, we have defined the filter weights for a CNN neuron. We can treat this neuron as a new feature that jointly predicts an object property along with the original neuron. If we find many decision boundaries in one visualization, it may be useful to first project the input feature space into 2D using the projection transformation we derived. Then we can implement a series of neurons that work on the 2d feature space, lowering the amount of parameters we need to store.

\subsection{Searching for semantics} 

The previous section describes an approach for uncovering the semantic concepts of individual neurons, as well as constructing new user-defined concepts. In theory, one could apply the given approach in an exhaustive fashion over all neighborhoods of neurons to uncover semantics of interest, but methods to filter the set of neurons can be used to reduce the search time.

We envision two modes of user interaction for our zero-shot pipeline. A user may have a semantic concept in mind ("I want to build a model of 3-wheeled vehicles"), or users may wish to peruse internal concepts captured in a network in an exploratory fashion. In either case, the user will identify a number of visual concepts which will then be assembled into a pictorial grammar (Sec.~\ref{sec:model}). In order to identify a pool of relevant candidates, a successful pipeline appeared to be first identifying an ``interesting" neuron (by random search at layers in {\tt conv5} or higher), and then searching for related neurons as detailed below. For example, after finding a "face" neuron, one might then search for neurons that co-activate, which might correspond to other body parts. Key to these methods is that statistics about the network and training set can be precomputed, making subsequent queries on the network available in constant time.

{\bf Filter search:} The first method utilizes the filter weights learned by the CNN. Recall that activations are computed by linear threshold functions \eqref{eq:activate} where the inputs $x$ themselves must be non-negative (since they are rectified activations themselves). This implies that the magnitude of weight of each input is directly related to the final activation value of a neuron, as well as its importance in positive feature detection for the neuron. Therefore, given a neuron of interest $a$, we search for connected neurons at higher and lower levels associated with large weights $w$. 
%we know that a particular neuron of interest will be activated when a neuron in the previous layer associated with a large positive weight
%Given a neuron of interest, we know that high-scoring patches should
%is based off of the learned filter weights. Each neuron determines the confidence of feature detection through a weighted sum of the neuron outputs from the previous layer. Because the ImageNet-CNN uses ReLU activations, all of the inputs to a neuron are strictly non-negative.

{\bf Co-occurrence search:} Our second method empirically searches for collections of neurons that consistently activate (over a set of images). Given a particular neuron of interest, we compute all non-zero activations on a validation set, and record the average activation of all connected neurons in higher and lower layers. We construct a pool of connected neurons (ordered by their filter weight or empirical correlation) that serve as candidates for embedding visualizations. We show examples of successfully-identified related neurons in Fig.~\ref{fig:search}.
%computes empirical correlations between the activations of a neuron of interest and overlapping neurons. This

%extracted over a validation set of images.

%activations across all layers across a validation set of images, and  pairs of neurons that  exhibit high correlations in their activations.
%The second method we use is based on the joint distribution of neuron activation between two layers in the network. 
%For the top 100-activations of a neuron, we collect the neuron inputs and average them across instances. The inputs that have a high average activation in this analysis correspond to features from the previous layer that are important to positive feature detection at the current neuron. 

%We show an application of this method in Fig \ref{fig:filter_weight_search}. Using these methods, we can quickly find a set of related neurons once we have found one interesting one.

%%%%%%%%%%%%%%%%%%%%%%%%%%%%%%%%%%%%%%%%%%%%%%%%%%%%%

%% Model Section

%%%%%%%%%%%%%%%%%%%%%%%%%%%%%%%%%%%%%%%%%%%%%%%%%%%%%

\section{Pictorial Language Classifiers}\label{sec:model}
\begin{figure*}[t]
    \centering
    \includegraphics[width=\linewidth]{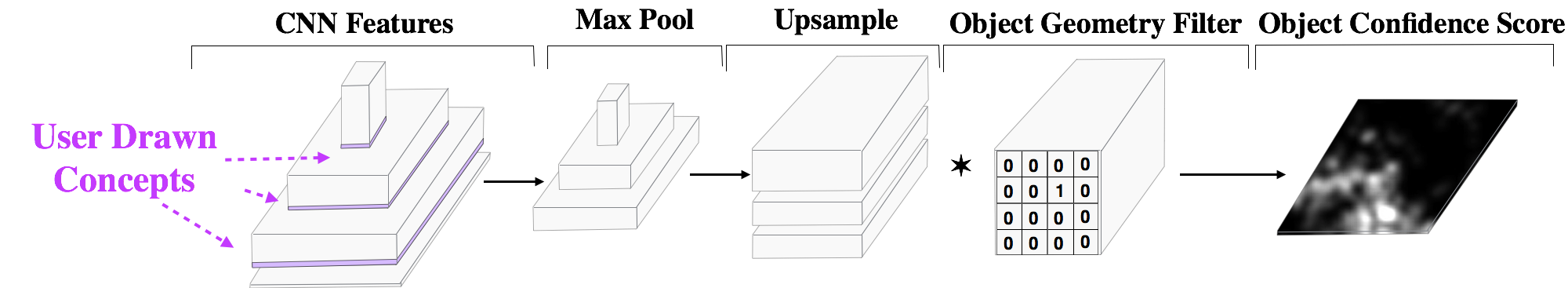}
    % \vspace{-10pt}
    \caption{%\textbf{Implementing pictorial grammars with CNNs}: Closely following the approach of \cite{girshick2015deformable},
    We efficiently implement pictorial grammars by augmenting pre-trained CNNs with additional user-defined concept filters (from Sec.~\ref{sec:viz}), max-pooling responses, upsampling responses across different layers to the same pixel resolution, and summing max-pooled response maps shifted by the anchor location of a part with respect to its parent (implemented as a sparse object geometry filter~\cite{girshick2015deformable}).}
%     \vspace{-10pt}%Note that we max-pool before upsampling to reduce computation. Though this procedure can be repeated by subsequent max-pools and convolutions, we visualize the pipeline for a "star model".}
%    with a few modifications. For our pipeline, we add a thresholding and sigmoid activation to normalize the feature maps produced by the CNN. Also, we remove the extra pooling operation and approximate a valid region in the object geometry filters.}
    \label{fig:dpm}
\end{figure*}

Our previous sections describe an interface for identifying visual tokens of interest. Examples might include textures ("striped"), part decompositions ("wheels"), and object invariances ("frontal-view faces"), that together can represent a wide range of complex objects. In this section, we construct a simple visual grammar for composing these tokens into flexible models for object detection. Importantly, the final visual grammar can be executed using standard CNNs (convolution and max-pooling) implying that the user-constructed models can be efficiently implemented as additional add-on layers to AlexNet. Since the vast majority of computation will be shared, this means that user-defined object detection essentially comes ``for-free".
%deformable part models for object detection using a combination of these CNN neurons and spatial AND-OR relationships. In addition, we formulate the entire pipeline using only convolutional operations, allowing our model to perform both ImageNet classification and object detection in one forward pass.

%\subsection{Model}
Our object detection model is based off the deformable parts model (DPMs) described in~\cite{felzenszwalb2010object}. In particular, we exploit the insight in \cite{girshick2015deformable} that DPMs can be implemented in CNN toolboxes. We revisit the derivation with slightly more detail here, pointing out that general tree-structured grammars can be implemented in CNNs. This allows, for example, users to build zero-shot detectors for objects (e.g., a ``striped car") as well as spatial arrangements of objects (e.g., ``a striped car next to two red bicycles"). %For ease of exposition, we focus on the "star-structured" variant that consists of root filter that covers the entire object, as well as a set of part filters representing localized parts. The feature set presented to the model can vary in resolution and robustness to small variances in inputs if the neurons selected for the model come from different layers of the network.
 Let us write the pixel location of part $i$ as $z_i = (x_i,y_i)$, and the score of a collection of parts $z = \{z_i\}$ as follows:
\begin{align}
    \text{score}(z) = \sum_{i} \phi(z_i) + \sum_{j \in parent(i)} \psi(z_j- z_i - a_j)
\end{align}
 The local function $\phi(z_i)$ denotes the score associated with placing part $i$ at pixel location $i$, which is found by evaluating (a possibly-upsampled) activation heatmap. % We also experiment with a thresholded activation map, based on the intuition that part scores may not be appropriately normalized.
 The pairwise function $\psi(z_j - z_i - a_j)$ encodes a spatial model that denotes valid relative locations of part $j$ to its parent $i$, where its rest anchor location is given by $a_j$. We use a simplified variant where 
 \begin{align}
     \psi(z_j - z_i - a_j) = 
     \begin{cases} 0, \quad \text{if} \quad ||z_j - z_i - a_j ||_\infty \leq r_j \\
     -\infty, \quad \text{otherwise}
     \end{cases}
 \end{align}
 which requires parts to lie within a square neighborhood (of width $2r_j$) of their anchor location.
 It is well-known that the best-scoring configuration of parts can be found with dynamic programming for tree-structured spatial constraints. By using the particular spatial model given above, partial scores of each part can be efficiently computed by repeating the following updates from the leaf to the root:
 %dynamic programming computations can be performed by max-pooling over $(2r+1) \times (2r+1)$ regions in each part's response map. Specifically, partial scores of each part given its children can be recursively computed by repeating the following, from leaf to root parts:
\begin{align*}
%\text{score}_i(z_i) = \phi(z_i) + \sum_{j \in kids(i)} \max_{\{z_j: ||z_j - z_i - a_j||_\infty \leq r_j\}} \text{score}_j(z_j) \label{eq:dp}
\text{score}_i(z_i) = \phi(z_i) + \sum_{j \in kids(i)} m_j(z_i + a_j), \quad \\\text{where} \quad m_j(z_i) = \max_{\{z_j: ||z_j - z_i||_\infty \leq r_j\}} \text{score}_j(z_j) 
\end{align*}
\\
Here $\text{score}_i(z_i)$ is initialized to $\phi(z_i)$ for leaf parts and $\text{score}_i(z_i)$ represents the true max-marginal score for the root part. Note that $m_j(z_i)$ is computed by max-pooling while $\text{score}_i(z_j)$ is computed by summing up shifted response maps (which can be implemented with a sparse, binary convolution operation). This implies that tree-structured dynamic programming operations can be implemented as standard layers in a CNN (Fig.~\ref{fig:dpm}).

{\bf User-defined spatial models:} Though a tree-structured formulation is flexible, it may be noninuitive for a user to manually specify. For that reason, we limit our results to "star" structured models that consist of a single root filter and a collection of child filters. We use an empty root filter of a fixed size $H \times W$, which specifies the bounding box dimension of a candidate detection. We allow users to specify the $a_j$ and $r_j$, which corresponds to the "rest" location of part within the bounding box, and the amount of valid displacements about that location. To find objects of different sizes, we process an image pyramid.

\section{Evaluation}
\label{sec:exp}

We evaluate our pictorial zero-shot framework with a variety of tasks of increasing difficulty, in terms of novelty with respect to AlexNet's training set. We first learn a model of categories that do appear in the training set but are {\em not labeled}. We then learn a {\em fine-grained} subcategory model for labeled categories, and finally conclude with a true zero-shot model of {\em never-before-seen} categories.

%\subsection{Unlabeled category: pedestrians}

\begin{figure*}[t!]
\centering
        \includegraphics[width=\linewidth]{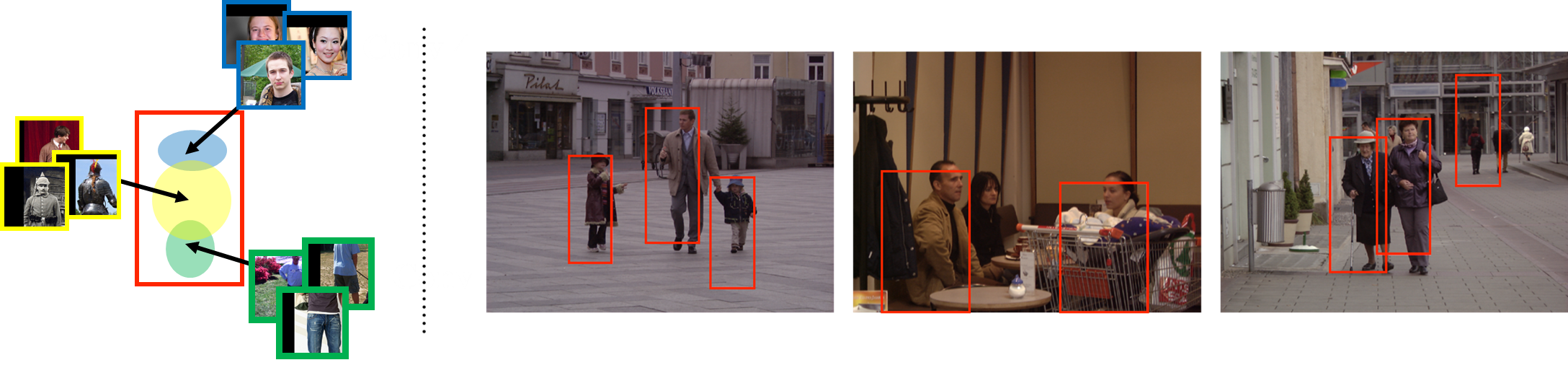}
%           \vspace{-10pt}
   \caption{On the \textbf{left}, we show a visual representation of the parts and spatial constraints used to construct our PLC for human detection. For each part, we draw an arrow to the spatial anchor in the template and visualize the allowed deformation of the part as a colored region. In practice, we use max pooling to enforce this deformation and thus the region will have a rectangular shape. At \textbf{right}, we show the top scoring detections for 3 instances of Graz-02. Quantitative results are included in Fig.~\ref{fig:human_ap}.}
%      \vspace{-10pt}
\label{fig:fig_examples}
\end{figure*}

{\bf Unlabeled category (Graz pedestrians):} Perhaps surprisingly, pedestrians are not an explicit category in the ILSVRC12's training set. We use our interface to construct a simple model for pedestrians using three user-defined parts, roughly corresponding to face, lower-body, and upper-body (visualized in Fig.~\ref{fig:fig_examples}).  We benchmark our model as a pedestrian detector on the Graz-02 \cite{marszatek2007accurate} dataset, which is composed of natural-scene images that contain cluttered backgrounds and complex objects in a variety of poses. We use the standard detection metric of average precision (AP) computed over various intersection-over-union (IOU) thresholds in Fig.~\ref{fig:human_ap}. We refer the reader to the caption for more details, but in summary, star-structured part models significantly outperform any single part, as well as a ``bag-of-parts" model without any spatial constraints. Our zero-shot models don't quite match the previously published performance of supervised models. One reason is that our models are tuned for frontal people. But our performance is impressive in that we use {\bf zero} labeled examples and {\bf no} linguistic knowledge!

\begin{figure}[t!]
\centering
        % Colour for the rulings in tables:
        %\makeatletter
        %   %\def\rulecolor#1#{\CT@arc{#1}}
        %   \def\CT@arc#1#2{%
        %   \ifdim\baselineskip=\z@\noalign\fi
        %   {\gdef\CT@arc@{\color#1{#2}}}}
        %   \let\CT@arc@\relax
        %  \rulecolor{gray!50}
        %\makeatother
%     \resizebox{\linewidth}{!}
    {
        \begin{tabular}{|c|c|c|c|}
        \hline
        Model & AP (0.3) & AP (0.5) & AP (0.8) \\
        %\midrule
        \hline
        PLC-Face & 0.152 & 0.123 & 0.0015 \\
        PLC-Lower Body & 0.412 & 0.036 & 0.006 \\
        PLC-Upper Body & 0.110 & 0.099 & 0.012 \\
        PLC-All (Bag)& 0.478 & 0.223 & 0.021 \\
        PLC-All (Spatial)& 0.696 & 0.499 & 0.110 \\
            \hline
%        \midrule
%        PLC-All with Thresholding & 0.568 & 0.489 & 0.061 \\
%        \midrule
%        \midrule
%        \midrule
        DPM \cite{felzenszwalb2010object} & - & 0.880 & - \\
        Fast R-CNN \cite{girshick2015fast} & - & 0.882 & - \\
    \hline  
        \end{tabular}}
%        {fig:fig_examples}.}
      \vspace{5pt}
   \caption{Evaluation of a human detector on Graz-O2. For each image, we run the detector (which is equivalent to running Alexnet convolutionally) at 5 scales and report the highest-scoring bounding boxes after NMS. We explore variants of a PLC with 3 parts, including individual parts, a bag-of-parts (without a spatial term), and a full 3-part model with star-spatial structure (which outperforms all variants). For comparison, we show average precision (AP) results of two supervised methods at varying overlap thresholds of .3, .5, and .8. While not state-of-the-art, our results are impressive given that no person labels were ever used. Our results are particularly accurate at lower overlap thresholds. Qualitative results are included in Fig.~\ref{fig:fig_examples}.}
%      \vspace{-10pt}
\label{fig:human_ap}
\end{figure}

\begin{figure}[ht!]
\centering
        \includegraphics[width=0.7\linewidth]{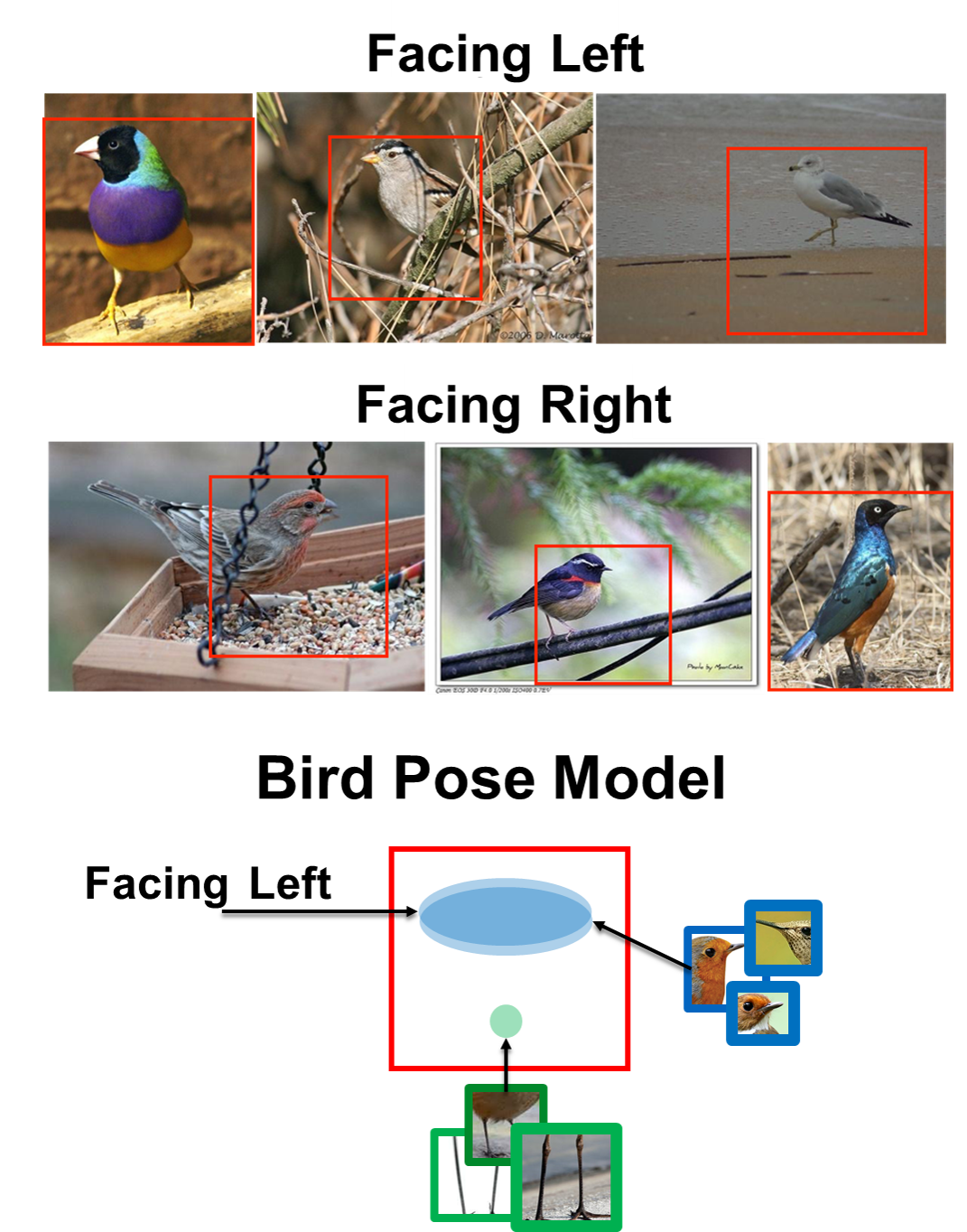}
   \caption{ We apply two {\em pose-specific} PLCs composed of 4 parts to detect birds and estimate their pose on PASCAL VOC 2007, showing the top-3 results for each detector. As in the previous visualization, we show both the anchor locations and allowable deformation for each set of parts. The part labeled \textit{facing left} refers to the bird pose neuron described in the text.  When we evaluate pose accuracy on the set of detected birds in PASCAL 2007-val, we obtain a correct classification rate of 87.5\% (where chance is roughly 50\%).} %\deva{Verify chance performance}.}
\label{fig:pascal}
\end{figure}

\begin{figure*}[t!]
\centering
        \includegraphics[width=\textwidth]{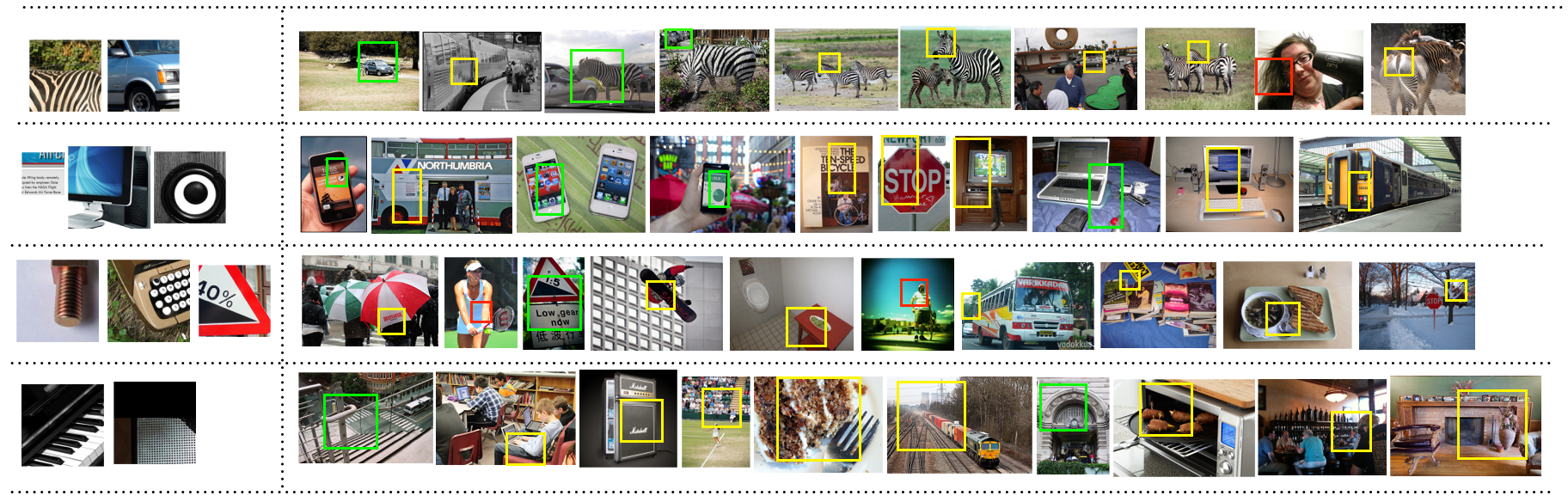}
%            \vspace{-10pt}
   \caption{Concept discovery on MSCOCO. Each row presents a set of neurons that composed into a ``bag-of-parts" PLC, followed by the top 10 detections from MSCOCO-val.  For reference, the visual concepts were originally selected by a user wishing to model a "striped car", a "cell phone", a "geometric shape with texture", and a purely random combination of neurons. We evaluate results with a user-study where participants (none of whom designed the queries) are prompted with the following: {\tt Interpret the query on the left as a set of visual concepts such as textures, parts, and objects. For each image on the right, determine whether the detection enclosed by the bounding box (i) represents all the queried visual concepts, (ii) represents at least one, or (iii) does not represent any.} Boxes are colored green (i), yellow (ii), or red (iii) based on the majority response. Precision for the four models at a recall of 10 is \text{0.401} for (i) and \text{0.931} for (ii), indicating that fairly accurate retrieval is possible even in the zero-shot setting.}
%       \vspace{-10pt}
\label{fig:coco}
\end{figure*}

\begin{figure*}[t!]
\centering
        \includegraphics[width=\linewidth]{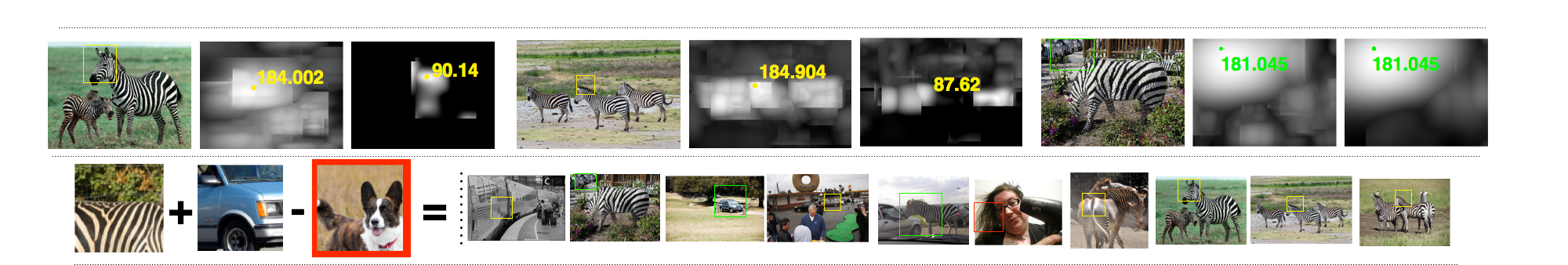}
%            \vspace{-10pt}
   \caption{We explore the notion of negative parts, by subtracting an {\em animal head} part score from a \textit{stripes}+\textit{car front}. On the \textbf{top}, we show three images corresponding to the query from the first row of Fig.~\ref{fig:coco}, followed by score maps of the original two-part model and the two-parts-minus-the-head. Scores associated with zebra regions decrease in magnitude, but the striped car detection (on the {\bf top-right}) remains high. This translates to a better re-ranking of retrieved images ({\bf bottom}).}%, we reorder the original 10 images using new PLC model. We see that the detections of zebra lower in score such they drop to the end of the retrieved set}
%       \vspace{-10pt}
\label{fig:minus}
\end{figure*}

{\bf Unlabeled subcategory (PASCAL birds):}
%As show in Section \ref{pca}, we can use PCA visualizations to manually create neurons that can discriminate between different object poses. Here, we use the decision boundary we determined in Fig \ref{fig:pca} to run a simple pose
We next learn novel subcategory for the labeled category of birds. We specifically use the user-drawn boundary in Fig.~\ref{fig:pca}-(a) to define left-facing and right-facing bird models. We evaluate our two pose-specific bird models on Pascal VOC 2007 testset, and show the top-3 hits for each model in Fig.~\ref{fig:pascal}. When we evaluate pose accuracy on the set of detected birds in PASCAL 2007-val, we obtain an accuracy of  87.5\%. %This implies that we can accurately estimate bird poses without any labeled examples. 
These results empirically support the claim that CNNs trained for image classification implicitly learn pose invariance, but this knowledge might be ``tangled" in internal neural activations. Our pictorial visualizations allow a user to ``untangle" such cues and explicitly build fine-grained detectors {\bf without} any labeled examples.

%Given a subclass, we can use our neuron search method to quickly identify the neurons that learn subclass pose invariance. Using this method, we can potentially build subclass pose detectors with only labeled classification data.

 %bird class. Specifically we construct two S, one for bird looking left and one for bird looking right, such that the pose of a bird is predicted by the max of the detection score outputed by each models. Fig \ref{fig:pascal} shows the top 3 scoring retrievals for each model; here the score not only reflects the confidence in bird orientation but also the confidence in bird detection. For the of bounding boxes correctly predicted for the bird subset of the PASCAL validation set, our pose detector achieves an accuracy of 87.5\%.

%
%\subsection{Concept Discovery on MS COCO}

{\bf Discovering concepts (MS COCO):} We now use zero-shot visualizations to discover novel concepts in MSCOCO~\cite{lin2014microsoft}. % We design a simple interface that allows  users to quickly specify a set of visual concepts, arrange them spatially in a template to construct a DPM, and perform queries on the COCO dataset for regions that maximally activate the model. 
 Fig.~\ref{fig:coco} shows 4 visual concept queries and the top 10 detections for each. For reference, the visual concepts were originally selected by a user wishing to model a "striped car", a "cell phone", a "geometric shape with texture", and a purely random combination of neurons. Each visual concept is user-defined to be a ``bag of (2-3) parts".
%each row shows the 10 images that maximally activate a model built with the set of visual concepts on the right. 
To evaluate results, we ask a set of participants whether or not the returned detections (i) encompass all of the visual parts, (ii) contain at least one of the visual parts, or (iii) are false detections. Our mAP for the four models at a recall of 10 is \text{0.401} for (i) and \text{0.931} for (ii). User-defined models tend to fire on image regions with a single strong part activation. We posit that since these parts rarely (if ever) co-occured in the training set, their activations are not properly calibrated, implying that better normalization of their activations might help. %We refer the reader to the caption for additional analysis, but visual concept discovery is surprisingly consistent. 
One interesting error mode is the \textit{striped car} query (first row), which actually fires on a striped car (the ranked-4 result) but also finds zebras. During our semantic exploration, it is quite apparent that an {\em animal head}s co-occur with {\em stripes}. We built a PLC with a "negative" part simply by subtracting the score of the best {\em animal head} within a candidate bounding box. We find that this leads to noticeably better score maps and retrievals (Fig.~\ref{fig:minus}), suggesting that negative parts might be a natural extension for making our interactive grammar more flexible.

%To solve this issue, we notice that the PLCs we have constructed so far have used parts in an \textit{excitatory} manner. Here, we notice that the visual concept \textit{animal head} is a characteristic of zebras but not striped cars; therefore we can use this concept in an \textit{inhibatory} hion to eliminate the false positives. Specifically, we build a PLC with the new part but importantly set the part weight to be negative. In Fig \ref{fig:minus}, we show that this negative part leads to a PLC that is a truly zero-shot classifier for striped cars

%better normalization of part scores could alleviate this problem.
%\section{Conclusion}

{\bf Conclusion:} We have shown that a CNN trained for ImageNet classification learns an image representation that can be transferred to build part-based detectors for (both previously seen but unlabeled and never-before-seen) objects without additional training data. To do so, we introduce the problem of {\em interactive zero-shot learning}, where a human-in-the-loop both (1) interprets and (2) reconfigures the internal semantics of a pre-trained neural net for such novel tasks. To (1) interpret internal semantics, we propose a novel perspective of convolutional neural nets as hierarchical compositions of {\em retinotopic embeddings}. We show that embedding visualizations can be used to interactively uncover semantic concepts that may not be captured in any single neuron. To (2) reconfigure these semantic concepts, we introduce a spatial grammar of {\em pictorial language classifiers} that make use of pictorial concepts rather than linguistic ones. Conveniently, pictorial language classifiers can be naturally integrated as "add-on" layers on top of exiting networks.
%also present a simple interface for users to intepret semantics associated with neighborhoods of neurons,
%and spatial constraints that can perform object detection, retrieval, and concept discovery on large datasets. A crucial aspect of our interface is a novel technique for visualizing local neighborhoods of neurons based on retinopic embeddings, which provide an alternative perspective of  %Our current interface implementation is not real-time, though we believe this is readily obtainable with intelligent feature caching (possible because our model can make use of off-the-shelf Alexnet features\\

\newpage
%\clearpage

%{\small
\bibliographystyle{ieee}
\bibliography{egbib}
%}

\end{document}